\documentclass[10pt,journal,compsoc]{IEEEtran}

\ifCLASSOPTIONcompsoc
\usepackage[nocompress]{cite}
\else
\usepackage{cite}
\fi

\ifCLASSINFOpdf
\usepackage[pdftex]{graphicx}
\else
\fi

\usepackage{amsmath}

\usepackage{url}

\usepackage{amsfonts} 
\usepackage{pgfplots}
\usepackage{subcaption}
\usepackage{makecell}
\usepackage{multirow}

\usepackage{lipsum}
\usepackage{placeins}
\usepackage{tabularx}

\usepackage{cleveref}
\usepackage{adjustbox}

\usepackage{array}

\usepackage{cite}
\newcommand\blfootnote[1]{%
	\begingroup
	\renewcommand\thefootnote{}\footnote{#1}%
	\addtocounter{footnote}{-1}%
	\endgroup
} 

\begin{document}
	
	\title{Federated Bayesian Network Ensembles}
	%
	%
	%
	%
	\author{Florian~van Daalen,~\IEEEmembership{Graduate Student Member,~IEEE,}
		Lianne~Ippel,
		Andre~Dekker,
		and~ Inigo~Bermejo
		\IEEEcompsocitemizethanks{\IEEEcompsocthanksitem  F. van Daalen, I. Bermejo, and A. Dekker are with 	the Department of Radiation Oncology (MAASTRO) GROW School for Oncology and Reproduction Maastricht University Medical Centre+ Maastricht the Netherlands\protect\\
			\IEEEcompsocthanksitem L. Ippel is with Statistics Netherlands Heerlen the Netherlands. }
	}
	
	
	\IEEEtitleabstractindextext{%
		\begin{abstract}
			Federated learning allows us to run machine learning algorithms on decentralized data when data sharing is not permitted due to privacy concerns. Ensemble-based learning works by training multiple (weak) classifiers whose output is aggregated. Federated ensembles are ensembles applied to a federated setting, where each classifier in the ensemble is trained on one data location. 
			In this article, we explore the use of federated ensembles of Bayesian networks (FBNE) in a range of experiments and compare their performance with locally trained models and models trained with VertiBayes, a federated learning algorithm to train Bayesian networks from decentralized data.  Our results show that FBNE outperforms local models and provides a significant increase in training speed compared with VertiBayes while maintaining a similar performance in most settings, among other advantages. We show that FBNE is a potentially useful tool within the federated learning toolbox, especially when local populations are heavily biased, or there is a strong imbalance in population size across parties. We discuss the advantages and disadvantages of this approach in terms of time complexity, model accuracy, privacy protection, and model interpretability.
		\end{abstract}
		
		}

	\maketitle

	\IEEEdisplaynontitleabstractindextext

	%
	
	\ifCLASSOPTIONcompsoc
	\IEEEraisesectionheading{\section{Introduction}\label{sec:introduction}}
	\else
	\section{Introduction}
	\label{sec:introduction}
	\fi
	
	\blfootnote{\emph{The views expressed in this paper are those of the authors and do not necessarily reflect the policy of Statistics Netherlands.}}
	\blfootnote{\emph{This research received funding from the Netherlands Organization for Scientific Research (NWO): Coronary ARtery disease: Risk estimations and Interventions for prevention and EaRly detection (CARRIER): project nr. 628.011.212.}}
	
	Federated learning allows machine learning algorithms to be applied to decentralized data when data sharing is not an option due to privacy concerns \cite{kairouz_advances_2021}. Traditionally federated learning approaches train a model iteratively on local data\cite{mcmahan_communication-efficient_2017}. The local results are then averaged back into a single global model. Privacy is preserved using epsilon-differential privacy\cite{dwork_algorithmic_2015}, homomorphic encryption\cite{parmar_survey_2014}, and multiparty computation (MPC)\cite{yao_protocols_1982} during this process. The specific techniques used, depend on the way the data is split across the various parties. If the data is split horizontally, i.e, each party has data belonging to a different population but the attributes are the same, simpler techniques can be used. While this approach yields good results in many cases, it suffers from several limitations.
	 
	A major downside is that it does not explicitly consider heterogeneity across different data sites. That is, it assumes that the data is independent and identically distributed (IID) over the various parties. However, in practice, federated environments will often be subject to local biases. For example, a common scenario where federated learning is implemented is when multiple hospitals combine their data to build a joint model\cite{deist_distributed_2020}. These hospitals may have very different population sizes, which may cause the final model to overfit on the biggest hospital. Additionally, the hospitals might have biases in their populations: e.g., an urban and a rural hospital will have different patient populations. Furthermore, the hospitals may even be on opposite sides of the globe, adding further biases into the distribution due to cultural, socio-economic, and many other factors. Simply averaging over these diverse populations may result in models that fit neither population, or it may result in the model overfitting on one particular population, while ignoring others.
	
	If the data is split vertically, , i.e. when different parties have different variables about the same population, it may be possible that there are dependencies between the parties. For example, if one party is a general practitioner (GP) and the other party is the specialist clinician, the GP might have referred the patient to the specialist and there will be a dependency between the two datasets. The GP might, for example, have started treatment based on the data he has, which will influence the data the specialist receives.
	
	These diverse types of bias may create problems when using the traditional federated learning approach. An alternative is the use of federated ensembles, ensembles of classifiers, each of which has been trained on the local data of each party in a federated setting\cite{van_daalen_federated_2022}. Ensemble based learning works by combining multiple (weak) classifiers which work together to jointly produce classifications using various voting schemes\cite{opitz_popular_1999}. It relies on a diverse set of classifiers, under the assumption that if one classifier makes a mistake the other classifiers will correct it. This allows ensembles to achieve a high performance, even when the individual classifiers are weak. A major advantage of ensemble learning is that it can deal with non-IID data\cite{van_daalen_ensemble_2018}. It can even take advantage of the dependencies by using (dynamically) weighted voting schemes. For example, an ensemble of experts can weigh the votes of classifiers trained on a similar population, or trained on specific sub-tasks, more strongly\cite{reisser_federated_2021, nguyen_distributed_2021}.
	
	Current research into federated ensembles is limited\cite{van_daalen_federated_2022}. There is only a small body of current work specifically looking into federated ensembles. There are still several general open questions, mainly:
	\begin{enumerate}
		\item How to share class labels in a privacy preserving manner in a vertically distributed setting.
		\item How to detect and exploit the subtle biases and dependencies that exist across parties.
		\item How to determine correlations between the various attributes split across parties.
	\end{enumerate}

	In addition to these general questions, there is also room for exploring different types of ensembles, using different base-classifiers. In this article, we will explore the use of federated ensembles consisting of Bayesian networks (BN). We will compare these ensembles with VertiBayes\cite{van_daalen_VertiBayes_2022}, a federated implementation of BNs, as well as with a BN that was centrally trained. We will compare the various options in terms of technical complexity, training time required, privacy protection, and model performance. 
	
	\section{Methods}
	\subsection{Bayesian networks}
	Bayesian networks (BN) are widely used probabilistic graphical models consisting of a directed acyclic graph (the structure) where each node represents a variables and arcs represent conditional dependencies and a set of conditional probability distributions (the parameters), one per node\cite{pearl_probabilistic_2014}. Their ability to combine existing expert knowledge with data has given them great utility and popularity. In addition, their graphical representation and probabilistic reasoning makes them relatively intuitive to understand models for non-technical personnel. This makes them especially useful in scenarios where non-technical personnel need to be able to understand the models, for example when models are used to inform clinical decisions.
	
	\subsection{VertiBayes}
	VertiBayes\cite{van_daalen_VertiBayes_2022} is an implementation of BN learning algorithms in a federated environment. It works both in vertical and horizontally distributed scenarios, as well as in a hybrid scenario. In addition to this, it can deal with missing data\cite{dempster_maximum_1977, lauritzen_em_1995}. Furthermore, it includes a federated implementation of the K2 algorithm\cite{cooper_bayesian_1992}, allowing it to learn a network structure on the fly. Lastly, VertiBayes includes several validation methods, of various computational complexity, that can be used to validate the model in a privacy preserving manner. This makes it an appropriate tool in a federated setting where data quality across parties may not be guaranteed.
	
	It has a similar performance compared to a centrally trained model. In addition, it provides the same privacy guarantees as the n-party scalar product protocol\cite{van_daalen_privacy_2023} used. However, it is considerably more time consuming to train a model using VertiBayes than it is to train a model centrally. The time complexity mostly depends on the number of probabilities that need to be calculated during parameter learning.

	\subsection{Federated Bayesian Network Ensembles}
	Federated Bayesian Network Ensembles (FBNE) are an ensemble learning approach where the base classifier consists of Bayesian networks. Each data owner within the federated setting makes their own Bayesian network based on locally available data. This local data is only enriched with the class label (should this not be available locally) in a privacy preserving manner using VertiBayes. The local classifiers can then be used in an ensemble to classify a new individual.
	 
	It is possible to use FBNE in a horizontally split, vertically split and hybrid settings. In the case of a hybrid split, one may decide to build the models purely based on local data, or to allow the hybrid variables to also utilize the data available at other data parties. For example, if party 1 contains attribute A \& B, and party 2 contains attributes B \& C, we can choose to either build a model using only the data available locally at party 1, or to build a model which also includes the data party 2 has regarding attribute B. In addition, it is possible with the use of predefined structures to mix and match variables from various parties to create the optimal ensemble. 	
	
	\subsubsection{Privacy risks} 
	The privacy risks posed by FBNE are the same as those posed by VertiBayes. That is to say, there are no major risks during training. However, the BNs themselves do still contain information. The structure and CPDs included in the BNs will be revealed if they are published. The published networks can be used to predict missing values in a dataset by a third party. This is inherent to how BNs work and as such is unavoidable. An ensemble of BNs poses a similar risk.
	
	\subsubsection{Runtime advantages}
	FBNE are significantly faster than VertiBayes as the majority of the calculations can be done locally. This minimizes the use of complex MPC needed to preserve privacy.
	
	\subsubsection{Performance advantages and disadvantages}
	FBNE may outperform a single model. The ensembles may catch local biases that are lost when only a single model is built. Furthermore, by using weighted voting it becomes possible to create a mixture of experts. This is especially advantageous if it is known that the various data parties have biases in their data. For example, in a scenario where hospitals work together to build an ensemble it may be useful to weigh the votes if one hospital specializes in a certain type of patient.
	
	\subsubsection{Interpretability}
	\label{Interpretability}
\begin{figure}
		\centering
		\begin{adjustbox}{width=\columnwidth}
			\includegraphics[scale=0.6]{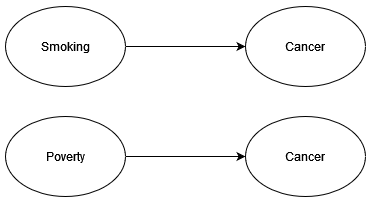}
		\end{adjustbox}
		\caption{Two local networks based on locally available data}
		\label{local}
	\end{figure}

	Ensembles are generally less interpretable than a single classifier. However, Bayesian networks are highly interpretable. While it is not possible to directly detect interactions across the various individual classifiers within the ensemble it is possible to use the individual classifiers to guide research into variable interactions in a smart manner. For example, by comparing the sparsity of the local networks to determine if certain variables are actually of interest to the outcome variable. In addition to this, it can be possible to use expert knowledge to deduce possible interactions.
	For example take the following toy examples shown in figure \ref{local}.
	
\begin{figure}
		\centering
		\begin{adjustbox}{width=\columnwidth}
			\includegraphics[scale=0.6]{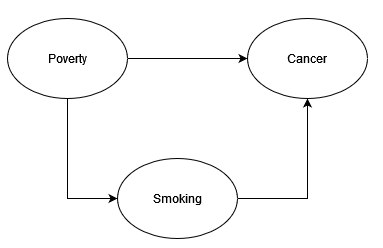}
		\end{adjustbox}
		\caption{Global network created by combining the two local models from figure \ref{local} while utilizing expert knowledge.}
		\label{global}
	\end{figure}

	In addition to these two models, expert knowledge indicates that poverty increases the likelihood of smoking. Based on this expert knowledge and the local models that were created it can be deduced that the global structure might be similar to the network shown in figure \ref{global}
	
	Furthermore, it should be noted that it is possible to apply feature selection across ensembles using wrapper-based approaches. By utilizing this type of feature selection, it may be possible to detect mediating effects between attributes.

	\section{Experiments} 
	We ran a range of experiments to assess the performance of FBNE\footnote{An implementation of FBNE can be found in the following git repository: \url{https://github.com/MaastrichtU-CDS/bayesianEnsemble}}. We used the following datasets:
	Iris\cite{de_marsico_mobile_2015}, Autism\cite{thabtah_autism_2017}, Asia\cite{lauritzen_local_1988}, Diabetes\cite{smith_using_1988}, Alarm\cite{beinlich_alarm_1989} and Mushroom\cite{schlimmer_concept_1987}.
	
	The Iris dataset contains 5 attributes and contains 150 individuals. It contains no missing values. The Autism dataset has 20 attributes and contains 704 individuals. The Asia dataset contains 10.000 individuals and 8 attributes.  The Alarm dataset also contains 10.000 individuals and 37 attributes. The Diabetes dataset contains 768 individuals and 9 attributes. The mushroom dataset has 23 attributes and contains 8124 individuals.
	
\begin{figure}
		\centering
		\begin{adjustbox}{width=\columnwidth}
			\includegraphics[scale=0.6]{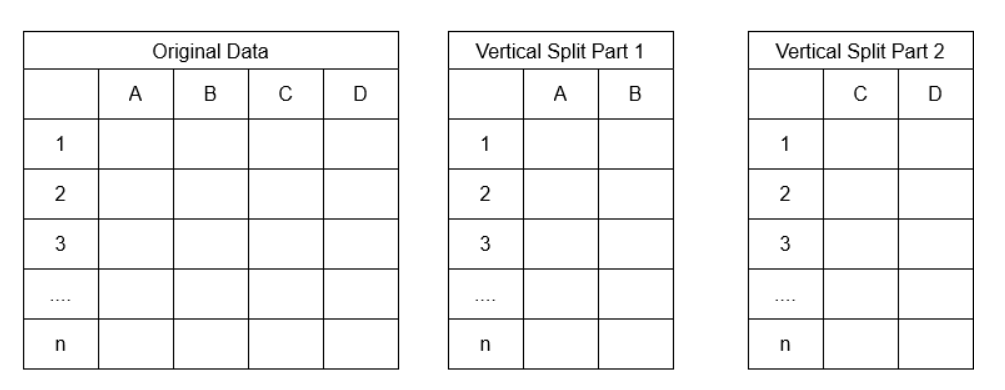}
		\end{adjustbox}
		\caption{Vertically split data}
		\label{vertical}
	\end{figure}

	Each dataset was tested with varying levels of missing values (no missing values, 5\%, 10\%, 30\% missing at random )
	We tested the following federated split scenarios:
	\begin{itemize}
		\item Vertical split with the attributes split randomly between parties. Each party has at least 2 local attributes. This was done for a setting with 2 and a setting with 3 parties. 
		\item Horizontal split with the samples split randomly between parties. Each party has at least 50 records. Again this was done for a setting with 2 and a setting with 3 parties.
		\begin{itemize}
			\item Additionally the horizontal splits were tested at varying levels of bias. To induce this bias we would make it more likely for individuals of a certain label to be put in a specific data station, e.g. individuals with the class label “true” are more likely to be put in party 1, individuals with class label “false” are more likely to be put in party 2. In the case of non-binary labels, such as for the Iris dataset, the bias would be created by piting label 1 against the rest. The following levels of bias were introduced: no bias, 75\%, 85\%, and 95\% bias. 
		\end{itemize}
		\item Hybrid split. The attributes are randomly split in two, the samples in one of these splits is then randomly split in two again. This results in 3 parties in total, each with at least 2 local attributes and 50 local records.   
	\end{itemize}
	An illustration of the various data splits can be found in \cref{horizontal,vertical,hybrid}.
	
	\begin{figure}
		\centering
		 \begin{adjustbox}{width=\columnwidth}
			\includegraphics[scale=0.6]{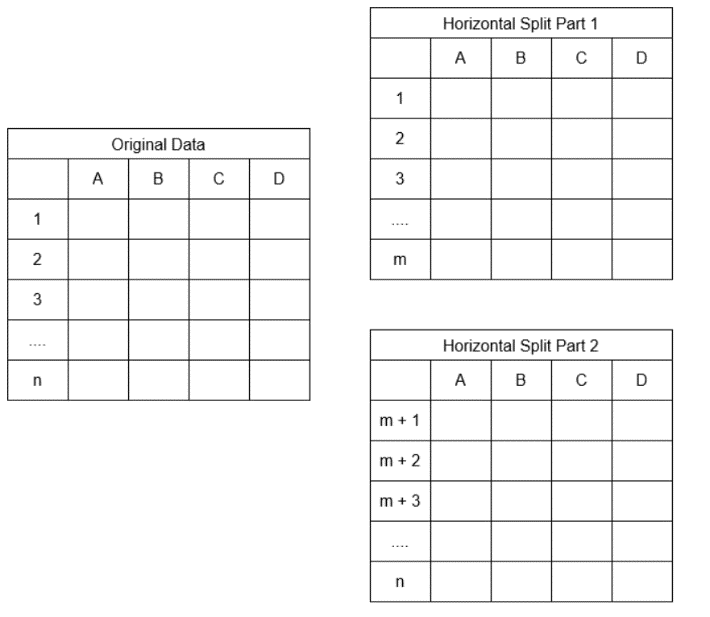}
		\end{adjustbox}
		\caption{Horizontally split data}
		\label{horizontal}
	\end{figure}	

	\begin{figure}
		\centering
		\begin{adjustbox}{width=\columnwidth}
			\includegraphics[scale=0.6]{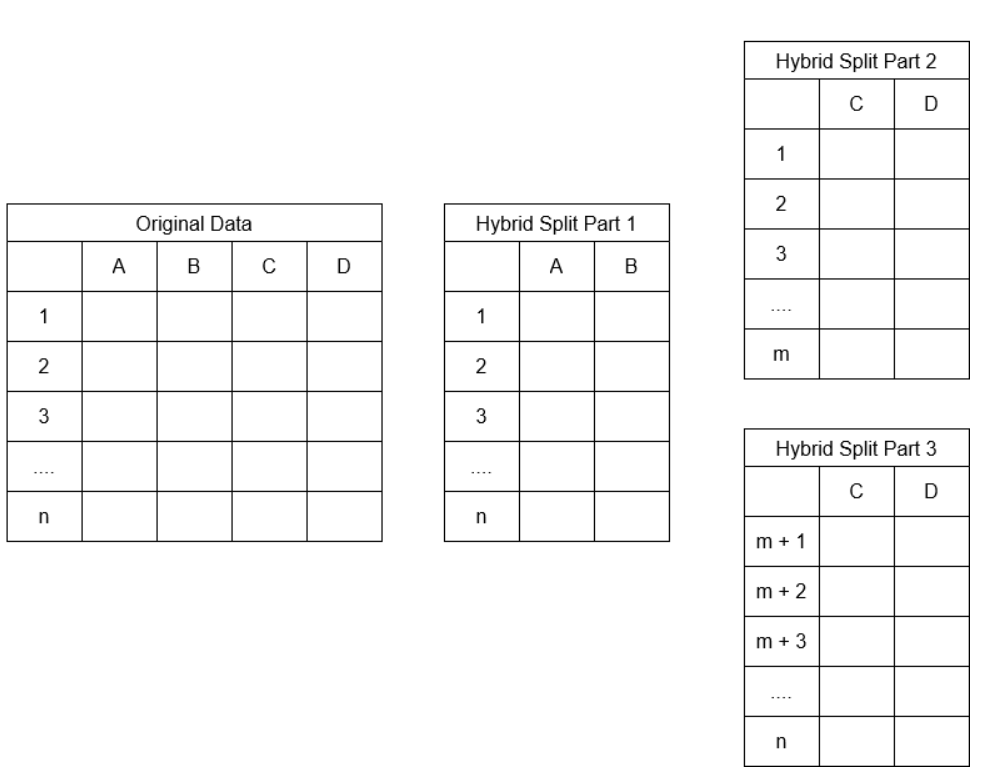}
		\end{adjustbox}
		\caption{Hybrid split data}
		\label{hybrid}
	\end{figure}	
		
	Each of these scenarios was run 10 times for each dataset. It should be noted that not every dataset could be used in every scenario. For example, the Iris dataset only contains 5 attributes and thus cannot be used in a 3-party vertically split scenario.
	
	In addition to these random splits, an experiment was also run where data was manually split in a “realistic” manner in a vertical setting based on expert knowledge. These experiments attempt to represent a realistic split where different parties collected different types of data. For example, the Autism data set contains attributes representing answers to a questionnaire, and some general personal attributes such as age, sex, and country of residence. This data was split in such a way that one party had the questionnaire answers, and the other party had the remaining attributes. As mentioned above, horizontal bias is addressed separately in the horizontal experiments.
	
	We compared the performance of FBNE with VertiBayes as well as a centrally trained Bayesian network. In addition, the performance of the local models on their local data is used as a baseline, after all if the local model already creates a sufficiently strong classifier there is no need for a federated model. The performance was measured by calculating the AUC using a 10-fold cross validation.
	
	In all cases both structure and parameters were learnt. Continuous variables were discretized into intervals that contained at least 10\% of the total population.	
	
	\section{Results}
	A selection of the results can be seen in tables \cref{vertical_automatic,vertical_biassed,hybrid,horizontal}. The selected results illustrate the general trends seen across all experiments. The remaining experimental results can be found in the appendix.
	 
	In all tables the highest AUC is indicated in green, the second highest in yellow.
	
	\begin{table*}[t]	
		\centering
		\caption{Experimental results vertically split 2-party scenarios where attributes were randomly split across parties. '*' Indicates the best performing model, '\dag' indicates the second best performing model.}
		\label{vertical_automatic}	
		\begin{tabular}{c|c|c|c|c|c|c}
		\multicolumn{2}{c}{} \vline & \multicolumn{5}{c}{AUC} \\ \hline													
		Name	&	Missing Data Level	&	FBNE	&	Party 1	&	Party 2	&	Central	&	VertiBayes	\\ \hline
		\makecell{Alarm \\ population size: 10000} 	&	0	&	0,888*	&	0,793\dag	&	0,675	&	0,789	&	0,789	\\ \hline 
		\multirow{4}{*}{\begin{tabular}{c}Asia \\ population size: 10000 \end{tabular}}	&	0	&	0,996*	&	0,918	&	0,886	&	0,995\dag	&	0,991	\\ 
		&	0.05	&	0,742\dag	&	0,694	&	0,696	&	0,735	&	0,776*	\\ 
		&	0.1	&	0,622\dag	&	0,594	&	0,591	&	0,615	&	0,677*	\\ 
		&	0.3	&	0,418	&	0,401	&	0,415	&	0,42\dag	&	0,581*	\\ \hline
		\multirow{4}{*}{\begin{tabular}{c}Autism \\ population size: 704 \end{tabular}}	&	0	&	0,909\dag	&	0,803	&	0,824	&	0,824	&	0,93*	\\ 
		&	0.05	&	0,8\dag	&	0,71	&	0,727	&	0,735	&	0,853*	\\ 
		&	0.1	&	0,737\dag	&	0,67	&	0,672	&	0,675	&	0,834*	\\ 
		&	0.3	&	0,531\dag	&	0,484	&	0,497	&	0,498	&	0,716*	\\ \hline
		\multirow{4}{*}{\begin{tabular}{c}Diabetes \\ population size: 768 \end{tabular}}	&	0	&	0,811*	&	0,744	&	0,69	&	0,78	&	0,808\dag	\\ 
		&	0.05	&	0,757\dag	&	0,658	&	0,684	&	0,723	&	0,772*	\\ 
		&	0.1	&	0,693\dag	&	0,597	&	0,624	&	0,668	&	0,767*	\\ 
		&	0.3	&	0,452\dag	&	0,418	&	0,406	&	0,44	&	0,667*	\\ \hline
		\multirow{4}{*}{\begin{tabular}{c}Iris \\ population size: 150 \end{tabular}}	&	0	&	0,94*	&	0,882	&	0,908\dag	&	0,885	&	0,75	\\ 
		&	0.05	&	0,892*	&	0,833	&	0,789	&	0,877\dag	&	0,787	\\ 
		&	0.1	&	0,782*	&	0,702	&	0,705	&	0,706\dag	&	0,701	\\ 
		&	0.3	&	0,658\dag	&	0,6	&	0,606	&	0,673*	&	0,607	\\ \hline 
		\makecell{Mushroom \\ population size: 8124}	&	0	&	0,999*	&	0,987\dag	&	0,898	&	0,999*	&	0,986	\\ \hline
		\end{tabular}
	\end{table*}

	\begin{table*}[t]
		\centering
		\caption{Experimental results vertically split 2-party scenarios where attributes were manually split across parties to simulate a realistic biased split. '*' Indicates the best performing model, '\dag' indicates the second best performing model.}
		\label{vertical_biassed}	
		\begin{tabular}{c|c|c|c|c|c|c}
			\multicolumn{2}{c}{} \vline & \multicolumn{5}{c}{AUC} \\ \hline													
			Name	&	Missing Data Level	&	FBNE	&	Party 1	&	Party 2	&	Central	&	VertiBayes	\\ \hline
			\multirow{4}{*}{\begin{tabular}{c}Autism \\ population size: 704 \end{tabular}}	&	0	&	0,889*	&	0,832	&	0,720	&	0,851\dag	&	0,834	\\
			&	0.05	&	0,789*	&	0,726	&	0,669	&	0,738	&	0,780\dag	\\
			&	0.1	&	0,728\dag	&	0,678	&	0,599	&	0,689	&	0,743*	\\
			&	0.3	&	0,497\dag	&	0,491	&	0,379	&	0,497\dag	&	0,625*	\\ \hline
			\multirow{4}{*}{\begin{tabular}{c}Iris \\ population size: 150 \end{tabular}}	&	0	&	0,940*	&	0,845	&	0,888	&	0,912\dag	&	0,748	\\
			&	0.05	&	0,887*	&	0,635	&	0,872	&	0,878\dag	&	0,736	\\
			&	0.1	&	0,774*	&	0,640	&	0,696	&	0,699\dag	&	0,671	\\
			&	0.3	&	0,654\dag	&	0,477	&	0,664	&	0,667*	&	0,622	\\ \hline
			\makecell{Mushroom \\ population size: 8124}	&	0	&	0,992*	&	0,880	&	0,987	&	0,987\dag	&	0,987\dag	\\ \hline	
		\end{tabular}
	\end{table*}

	\begin{table*}[t]
		\label{hybrid}
		\centering
		\caption{Experimental results hybrid split 3-party scenarios where only locally available data was used in the local model. '*' Indicates the best performing model, '\dag' indicates the second best performing model.}
		\begin{tabular}{c|c|c|c|c|c|c|c|c}
			\multicolumn{2}{c}{} \vline & \multicolumn{6}{c}{AUC} \\ \hline													
			Name	&	Missing Data Level	&	FBNE	&	Party 1	&	Party 2	 & Party 3 &	Central	&	VertiBayes	\\ \hline
			\multirow{4}{*}{\begin{tabular}{c}Asia \\ population size: 10000 \end{tabular}}	&	0	&	0,996*	&	0,928	&	0,940	&	0,938	&	0,995\dag	&	0,987	\\
			&	0.05	&	0,742	&	0,689	&	0,670	&	0,670	&	0,746\dag	&	0,789*	\\
			&	0.1	&	0,627\dag	&	0,611	&	0,596	&	0,595	&	0,624	&	0,668*	\\
			&	0.3	&	0,418\dag	&	0,400	&	0,406	&	0,407	&	0,418\dag	&	0,568*	\\ \hline
			\multirow{4}{*}{\begin{tabular}{c}Autism \\ population size: 704 \end{tabular}}	&	0	&	0,911*	&	0,805	&	0,821	&	0,820	&	0,833\dag	&	0,826	\\
			&	0.05	&	0,803\dag	&	0,707	&	0,718	&	0,719	&	0,741	&	0,932*	\\
			&	0.1	&	0,739\dag	&	0,671	&	0,667	&	0,665	&	0,698	&	0,891*	\\
			&	0.3	&	0,537\dag	&	0,473	&	0,493	&	0,493	&	0,491	&	0,783*	\\ \hline
			\multirow{4}{*}{\begin{tabular}{c}Diabetes \\ population size: 768 \end{tabular}}	&	0	&	0,814*	&	0,700	&	0,734	&	0,735	&	0,776\dag	&	0,781	\\
			&	0.05	&	0,749\dag	&	0,685	&	0,657	&	0,656	&	0,726	&	0,753*	\\
			&	0.1	&	0,694\dag	&	0,633	&	0,596	&	0,607	&	0,676	&	0,740*	\\
			&	0.3	&	0,452\dag	&	0,420	&	0,407	&	0,409	&	0,440	&	0,691*	\\ \hline
			\multirow{4}{*}{\begin{tabular}{c}Iris \\ population size: 150 \end{tabular}}	&	0	&	0,952*	&	0,912\dag	&	0,888	&	0,894	&	0,885	&	0,700	\\
			&	0.05	&	0,886*	&	0,784	&	0,849	&	0,853	&	0,874\dag	&	0,805	\\
			&	0.1	&	0,798*	&	0,725	&	0,730	&	0,725	&	0,688	&	0,642	\\
			&	0.3	&	0,633\dag	&	0,603	&	0,575	&	0,582	&	0,669*	&	0,610	\\ \hline
			
		\end{tabular}
	\end{table*}

	\begin{table*}[t]
		\label{horizontal}
		\centering
		\caption{Experimental results horizontally split 2-party scenarios where records are randomly split across parties. '*' Indicates the best performing model, '\dag' indicates the second best performing model.}
		\begin{tabular}{c|c|c|c|c|c|c}
			\multicolumn{2}{c}{} \vline & \multicolumn{5}{c}{AUC} \\ \hline													
			Name	&	Missing Data Level	&	FBNE	&	Party 1	&	Party 2	&	Central	&	VertiBayes	\\ \hline
			\multirow{4}{*}{\begin{tabular}{c}Asia \\ population size: 10000 \end{tabular}}	&	0	&	0,996*	&	0,995\dag	&	0,995\dag	&	0,995\dag	&	0,988	\\
			&	0.05	&	0,740	&	0,741\dag	&	0,741\dag	&	0,740	&	0,765*	\\
			&	0.1	&	0,624\dag	&	0,623	&	0,623	&	0,624\dag	&	0,670*	\\
			&	0.3	&	0,416	&	0,418\dag	&	0,418\dag	&	0,415	&	0,568*	\\ \hline
			\multirow{4}{*}{\begin{tabular}{c}Autism \\ population size: 704 \end{tabular}}	&	0	&	0,868*	&	0,777	&	0,774	&	0,847\dag	&	0,834	\\
			&	0.05	&	0,787*	&	0,464	&	0,464	&	0,737	&	0,781\dag	\\
			&	0.1	&	0,691\dag	&	0,466	&	0,463	&	0,691\dag	&	0,746*	\\
			&	0.3	&	0,534\dag	&	0,392	&	0,413	&	0,492	&	0,628*	\\ \hline
			\multirow{4}{*}{\begin{tabular}{c}Diabetes \\ population size: 768 \end{tabular}}	&	0	&	0,781\dag	&	0,500	&	0,500	&	0,781\dag	&	0,782*	\\
			&	0.05	&	0,725	&	0,480	&	0,480	&	0,729\dag	&	0,752*	\\
			&	0.1	&	0,671	&	0,447	&	0,447	&	0,680\dag	&	0,736*	\\
			&	0.3	&	0,441\dag	&	0,431	&	0,437	&	0,437	&	0,648*	\\ \hline
			\multirow{4}{*}{\begin{tabular}{c}Iris \\ population size: 150 \end{tabular}}	&	0	&	0,957*	&	0,903	&	0,898	&	0,925\dag	&	0,782	\\
			&	0.05	&	0,877\dag	&	0,829	&	0,837	&	0,879*	&	0,759	\\
			&	0.1	&	0,790*	&	0,749	&	0,774\dag	&	0,7049566	&	0,675	\\
			&	0.3	&	0,611\dag	&	0,494	&	0,533	&	0,677*	&	0,609	\\ \hline
			\makecell{Mushroom \\ population size: 8124}	&	0	&	0,988*	&	0,988*	&	0,988*	&	0,987\dag	&	0,987\dag	\\ \hline
		\end{tabular}
	\end{table*}
	
	\subsection{Runtime}
	The training process for FBNE is consistently faster than VertiBayes in every experimental setting and for every dataset used. However, it should be noted that the differences vary widely and depend on the dataset. These differences are mainly driven by the different network structures. FBNE has the advantage that most calculations can be done locally. During our experiments FBNE was faster than VertiBayes by a factor that ranged from twice as fast to fifty times as fast. It should be noted that it is difficult to predict per scenario how large the processing speed gains will be as it is difficult to predict the resulting network structure.
	
	\subsection{Performance}
	FBNE largely shows the same performance as VertiBayes, scoring similar AUC’s. However, it should be noted that for specific datasets, and for specific data-splits, the ensembles can perform significantly better. With the largest difference being nearly a 0.1 difference in AUC.
	 
	In addition to outperforming VertiBayes by a relevant margin in specific scenarios, the following general trends were visible in our experiments. First, FBNE performs very well in scenarios with no missing data. FBNE achieved the highest AUC in $80\%$ of the scenarios with no missing data. However, VertiBayes performs better in scenarios with missing values. VertiBayes achieved the highest AUC in roughly $70\%$ of the scenarios.
	
	Another interesting trend that is visible in our experiments is that in horizontally split scenarios the local models can perform well if the local data quality is high, occasionally performing similarly to the federated and centralized models. However, it should be noted that this is rare.
	
	It is important to note that neither approach was optimized, and better models can potentially be created. For example, a network structure generated using expert knowledge, as opposed to using an automatic approach like the K2 algorithm, may result in better Bayesian networks. Both VertiBayes and FBNE could benefit from such expert knowledge. In addition, weighted voting, and especially dynamically weighted voting, can improve the performance of FBNE.
	
	\section{Discussion}
	Our experiments show that FBNE can be a suitable solution in certain scenarios. In this section we will take a deeper dive into the differences between FBNE and VertiBayes.
	
	\subsection{Runtime}
	The reduced training time that was observed during the experiments is a strong advantage in favor of the federated ensembles, especially in time critical applications or in cases when MPC solutions such as VertiBayes are too time consuming. FBNE has this advantage due to the fact that the vast majority of calculations can be done locally and thus requires far fewer computationally difficult operations than VertiBayes does.
	
	In addition to the improved runtime already observed here, it should be noted that our implementation of the ensembles has not been fully optimized. There is room for further improvements, especially with respect to parallelization, which will further increase the gap in terms of runtime. However, since the goal of this study was to simply explore the potential of FBNE, not to provide an optimized implementation, this will not a part of this study.
	
	\subsection{Performance}
	
	FBNE largely showed the same performance as VertiBayes, however, in specific scenarios it significantly outperformed VertiBayes. With the largest difference being nearly a 0.1 difference in AUC. This does indicate that FBNE are potentially very useful in the right situation. However, it is very difficult to determine when this is the case without simply training the FBNE.
	
	Two trends were visible with respect to the performance differences. First, FBNE performs very well in scenarios with no missing data. However, VertiBayes performs better in scenarios with missing values. There are two plausible reasons that explain why VertiBayes performs better in these scenarios. The first explanation is that FBNE did not have a large, or diverse, enough ensemble in the experimental scenarios to work properly. Ensemble learning relies on a diverse set of classifiers which can correct each other’s mistakes, with only 2 or 3 models in our scenarios the ensembles may not be able to do this consistently. The second possible explanation is that the synthetic data generation step within VertiBayes allows it to bootstrap itself for an improved performance.
	 
	Another interesting trend that is visible in our experiments is that in horizontally split scenarios the local models occasionally perform well when local data is of a high quality, especially when the local population is not biased in any way. This reminds us that it is always important to ask if a federated model is truly necessary. Building a federated model is only worthwhile if the data added from other parties adds extra information. But if your local data is already sufficiently large, and representative of the true population, then a federated model may not be needed. However, if the local data is small, or biased in some way, then a federated approach is needed.
	 
	It is important to note that neither approach was optimized, and better models can potentially be created. For example, both VertiBayes and FBNE can benefit from improvements, such as using expert knowledge to build the optimal structures. In addition to improvements that could be applied to both approaches, weighted voting, and especially dynamically weighted voting, can improve the performance of FBNE.

	\subsection{Privacy concerns \& disclosure control}
	Our implementation of FBNE uses VertiBayes at its core. As such, the privacy guarantees are largely the same. However, there are two aspects in which they potentially differ from VertiBayes:
	\begin{enumerate}
		\item The classification of individuals, and evaluation of the model.
		\item The consequences of having multiple networks.
	\end{enumerate} 
	
	With FBNE, the individual classifiers can create their classifications fully locally at the party they belong to given that required attributes for each local model should be available locally. These individual classifications can be combined using homomorphic encryption, resulting in a final classification which can be shared. For example, estimated probabilities can be weighted according to the voting power of a particular classifier, then encrypted using an additive homomorphic encryption scheme, and all encrypted weighted probabilities are summed. The sum is then decrypted and divided by the total weight to get the weighted average probabilities, which determine the final classification. Combining the votes in this way prevents any local data from being shared and allows FBNE to be evaluated without the need of the homomorphic encryption and a privacy preserving n-scalar product protocol\cite{van_daalen_privacy_2023} or other more complex evaluation methods that VertiBayes needs \cite{van_daalen_vertibayes_2022}. This is a strong advantage when new samples need to be classified or predicted in a federated manner.
	
	The other aspect in which FBNE differs from VertiBayes is that the end-result consists of multiple networks instead of one. Ensembles might provide a minor advantage with respect to privacy in this case. In both cases it is possible to learn dependencies between attributes and certain statistics about the training set, based on the CPDs and network structures. Similarly, based on local, incomplete data, the network can be used to predict missing values. These are unavoidable consequences of using Bayesian networks. However, because FBNE has the information split up over multiple networks, it will be difficult to do this for every attribute. It is very difficult to determine any relation between two attributes when those two attributes are split over two Bayesian networks, as discussed in section \ref{Interpretability}. This provides some additional privacy protection compared to a single network. 
	
	\subsection{When should FBNE be preferred over VertiBayes}
	Due to the significant advantage in terms of runtime it may be beneficial to use FBNE as an exploratory first step before deciding if using VertiBayes is worth it. This naïve approach will already provide reasonable results.
	
	The general advantages of each approach can be found in table \ref{comparison}.
		
	\begin{table}[t]
		\centering
		\caption{Comparison of main features of FBNE and VertiBayes}
		\label{comparison}	
		\begin{tabularx}{\columnwidth}{X|X}
			FBNE & VertiBayes \\ \hline
			Faster & More complete view of dependencies between attributes \\ \hline
			Slightly better privacy guarantees & Easier to understand \& interpret than an ensemble \\ \hline
			Possible to capture biases in local population & May outperform FBNE when the ensemble is too small or not diverse enough \\ \hline
			Can easily classify new samples in a privacy preserving manner & \\ 
		\end{tabularx}
	\end{table}
	
	In addition to these advantages which hold in general, one of the two may perform better depending on how the data happened to be split. However, there is currently no good way to predict which approach will achieve the highest accuracy.
	
	\section{Conclusion}
	In this article, we have proposed the use of Federated Bayesian Network Ensembles (FBNE) and assessed their usefulness in a battery of experiments. We have shown the approach performs well in a range of situations and datasets, often achieving similar results when compared to VertiBayes, an alternative federated method. FBNE are significantly faster than VertiBayes, provide slightly better privacy guarantees, and are easier to use in a scenario where future classifications will also be done in a federated setting. On the other hand, VertiBayes results in more interpretable models and makes it easier to determine the dependencies between variables split over multiple sites.
	
	The notable advantage in terms of runtime does mean that it is easily possible to use FBNE as an initial exploratory option. Since it is currently not possible to preemptively determine which approach will result in the highest accuracy, using this naive approach and simply training both models might be the best course of action for now. Additionally, this means it can be very useful when exploring new federated datasets.

	\subsection{Future work}
	We would like to explore ways to determine which approach is more effective as it would be highly beneficial to be able to know beforehand if an ensemble-based approach will outperform a single model. Additionally, we would like to run experiments to discover if (dynamically) weighted voting could be used to significantly improve the performance of the ensembles. Lastly, it would be extremely valuable if these experiments could be run on real use cases. This would allow the experiments to work with realistic biases and remove the need to artificially create these biases in our experimental setup. Resulting in much more realistic experimental scenarios. 
	
	\bibliographystyle{IEEETran}
	\bibliography{FBNE}

\begin{thebibliography}{10}
\providecommand{\url}[1]{#1}
\csname url@samestyle\endcsname
\providecommand{\newblock}{\relax}
\providecommand{\bibinfo}[2]{#2}
\providecommand{\BIBentrySTDinterwordspacing}{\spaceskip=0pt\relax}
\providecommand{\BIBentryALTinterwordstretchfactor}{4}
\providecommand{\BIBentryALTinterwordspacing}{\spaceskip=\fontdimen2\font plus
\BIBentryALTinterwordstretchfactor\fontdimen3\font minus
  \fontdimen4\font\relax}
\providecommand{\BIBforeignlanguage}[2]{{%
\expandafter\ifx\csname l@#1\endcsname\relax
\typeout{** WARNING: IEEEtran.bst: No hyphenation pattern has been}%
\typeout{** loaded for the language `#1'. Using the pattern for}%
\typeout{** the default language instead.}%
\else
\language=\csname l@#1\endcsname
\fi
#2}}
\providecommand{\BIBdecl}{\relax}
\BIBdecl

\bibitem{kairouz_advances_2021}
\BIBentryALTinterwordspacing
P.~Kairouz, H.~B. McMahan, B.~Avent, A.~Bellet, M.~Bennis, A.~N. Bhagoji,
  K.~Bonawitz, Z.~Charles, G.~Cormode, R.~Cummings, R.~G.~L. D’Oliveira,
  H.~Eichner, S.~E. Rouayheb, D.~Evans, J.~Gardner, Z.~Garrett, A.~Gascón,
  B.~Ghazi, P.~B. Gibbons, M.~Gruteser, Z.~Harchaoui, C.~He, L.~He, Z.~Huo,
  B.~Hutchinson, J.~Hsu, M.~Jaggi, T.~Javidi, G.~Joshi, M.~Khodak, J.~Konecný,
  A.~Korolova, F.~Koushanfar, S.~Koyejo, T.~Lepoint, Y.~Liu, P.~Mittal,
  M.~Mohri, R.~Nock, A.~Özgür, R.~Pagh, H.~Qi, D.~Ramage, R.~Raskar,
  M.~Raykova, D.~Song, W.~Song, S.~U. Stich, Z.~Sun, A.~T. Suresh, F.~Tramèr,
  P.~Vepakomma, J.~Wang, L.~Xiong, Z.~Xu, Q.~Yang, F.~X. Yu, H.~Yu, and
  S.~Zhao, ``\BIBforeignlanguage{English}{Advances and {Open} {Problems} in
  {Federated} {Learning}},'' \emph{\BIBforeignlanguage{English}{Foundations and
  Trends® in Machine Learning}}, vol.~14, no. 1–2, pp. 1--210, Jun. 2021,
  publisher: Now Publishers, Inc. [Online]. Available:
  \url{https://www.nowpublishers.com/article/Details/MAL-083}
\BIBentrySTDinterwordspacing

\bibitem{mcmahan_communication-efficient_2017}
\BIBentryALTinterwordspacing
B.~McMahan, E.~Moore, D.~Ramage, S.~Hampson, and B.~A.~y. Arcas,
  ``\BIBforeignlanguage{en}{Communication-{Efficient} {Learning} of {Deep}
  {Networks} from {Decentralized} {Data}},'' in
  \emph{\BIBforeignlanguage{en}{Proceedings of the 20th {International}
  {Conference} on {Artificial} {Intelligence} and {Statistics}}}.\hskip 1em
  plus 0.5em minus 0.4em\relax PMLR, Apr. 2017, pp. 1273--1282, iSSN:
  2640-3498. [Online]. Available:
  \url{https://proceedings.mlr.press/v54/mcmahan17a.html}
\BIBentrySTDinterwordspacing

\bibitem{dwork_algorithmic_2015}
\BIBentryALTinterwordspacing
C.~Dwork and A.~Roth, ``\BIBforeignlanguage{en}{The {Algorithmic} {Foundations}
  of {Differential} {Privacy}},'' \emph{\BIBforeignlanguage{en}{Foundations and
  Trends® in Theoretical Computer Science}}, vol.~9, no. 3-4, pp. 211--407,
  Aug. 2015. [Online]. Available:
  \url{http://www.nowpublishers.com/articles/foundations-and-trends-in-theoretical-computer-science/TCS-042}
\BIBentrySTDinterwordspacing

\bibitem{parmar_survey_2014}
\BIBentryALTinterwordspacing
P.~V. Parmar, S.~B. Padhar, S.~N. Patel, N.~I. Bhatt, and R.~H. Jhaveri,
  ``\BIBforeignlanguage{en}{Survey of {Various} {Homomorphic} {Encryption}
  algorithms and {Schemes}},'' \emph{\BIBforeignlanguage{en}{International
  Journal of Computer Applications}}, vol.~91, no.~8, pp. 26--32, Apr. 2014.
  [Online]. Available:
  \url{http://research.ijcaonline.org/volume91/number8/pxc3895081.pdf}
\BIBentrySTDinterwordspacing

\bibitem{yao_protocols_1982}
A.~C. Yao, ``Protocols for secure computations,'' in \emph{23rd {Annual}
  {Symposium} on {Foundations} of {Computer} {Science} (sfcs 1982)}, Nov. 1982,
  pp. 160--164, iSSN: 0272-5428.

\bibitem{deist_distributed_2020}
\BIBentryALTinterwordspacing
T.~M. Deist, F.~J. W.~M. Dankers, P.~Ojha, M.~Scott~Marshall, T.~Janssen,
  C.~Faivre-Finn, C.~Masciocchi, V.~Valentini, J.~Wang, J.~Chen, Z.~Zhang,
  E.~Spezi, M.~Button, J.~Jan~Nuyttens, R.~Vernhout, J.~van Soest, A.~Jochems,
  R.~Monshouwer, J.~Bussink, G.~Price, P.~Lambin, and A.~Dekker,
  ``\BIBforeignlanguage{en}{Distributed learning on 20 000+ lung cancer
  patients – {The} {Personal} {Health} {Train}},''
  \emph{\BIBforeignlanguage{en}{Radiotherapy and Oncology}}, vol. 144, pp.
  189--200, Mar. 2020. [Online]. Available:
  \url{https://www.sciencedirect.com/science/article/pii/S0167814019334899}
\BIBentrySTDinterwordspacing

\bibitem{van_daalen_federated_2022}
\BIBentryALTinterwordspacing
F.~van Daalen, L.~Ippel, A.~Dekker, and I.~Bermejo,
  ``\BIBforeignlanguage{en}{Federated {Ensembles}: a literature review},'' Dec.
  2022. [Online]. Available: \url{https://www.researchsquare.com}
\BIBentrySTDinterwordspacing

\bibitem{opitz_popular_1999}
\BIBentryALTinterwordspacing
D.~Opitz and R.~Maclin, ``\BIBforeignlanguage{en}{Popular {Ensemble} {Methods}:
  {An} {Empirical} {Study}},'' \emph{\BIBforeignlanguage{en}{Journal of
  Artificial Intelligence Research}}, vol.~11, pp. 169--198, Aug. 1999.
  [Online]. Available: \url{https://jair.org/index.php/jair/article/view/10239}
\BIBentrySTDinterwordspacing

\bibitem{van_daalen_ensemble_2018}
F.~van Daalen, E.~Smirnov, N.~Davarzani, R.~Peeters, J.~Karel, and H.~B.-L.
  Rocca, ``An {Ensemble} {Approach} to {Time} {Dependent} {Classification},''
  in \emph{2018 17th {IEEE} {International} {Conference} on {Machine}
  {Learning} and {Applications} ({ICMLA})}, Dec. 2018, pp. 1007--1011.

\bibitem{reisser_federated_2021}
\BIBentryALTinterwordspacing
M.~Reisser, C.~Louizos, E.~Gavves, and M.~Welling, ``Federated {Mixture} of
  {Experts},'' p. arXiv:2107.06724, Jul. 2021. [Online]. Available:
  \url{https://ui.adsabs.harvard.edu/abs/2021arXiv210706724R}
\BIBentrySTDinterwordspacing

\bibitem{nguyen_distributed_2021}
\BIBentryALTinterwordspacing
M.~N.~H. Nguyen, S.~R. Pandey, K.~Thar, N.~H. Tran, M.~Chen, W.~S. Bradley, and
  C.~S. Hong, ``Distributed and {Democratized} {Learning}: {Philosophy} and
  {Research} {Challenges},'' \emph{IEEE Computational Intelligence Magazine},
  vol.~16, no.~1, pp. 49--62, 2021. [Online]. Available:
  \url{https://ieeexplore.ieee.org/document/9321418/}
\BIBentrySTDinterwordspacing

\bibitem{van_daalen_VertiBayes_2022}
\BIBentryALTinterwordspacing
F.~van Daalen, L.~Ippel, A.~Dekker, and I.~Bermejo, ``{VertiBayes}: {Learning}
  {Bayesian} network parameters from vertically partitioned data with missing
  values,'' Oct. 2022, arXiv:2210.17228 [cs]. [Online]. Available:
  \url{http://arxiv.org/abs/2210.17228}
\BIBentrySTDinterwordspacing

\bibitem{pearl_probabilistic_2014}
J.~Pearl, \emph{\BIBforeignlanguage{en}{Probabilistic {Reasoning} in
  {Intelligent} {Systems}: {Networks} of {Plausible} {Inference}}}.\hskip 1em
  plus 0.5em minus 0.4em\relax Elsevier, Jun. 2014, google-Books-ID:
  mn2jBQAAQBAJ.

\bibitem{dempster_maximum_1977}
\BIBentryALTinterwordspacing
A.~P. Dempster, N.~M. Laird, and D.~B. Rubin, ``Maximum {Likelihood} from
  {Incomplete} {Data} via the {EM} {Algorithm},'' \emph{Journal of the Royal
  Statistical Society. Series B (Methodological)}, vol.~39, no.~1, pp. 1--38,
  1977, publisher: [Royal Statistical Society, Wiley]. [Online]. Available:
  \url{https://www.jstor.org/stable/2984875}
\BIBentrySTDinterwordspacing

\bibitem{lauritzen_em_1995}
\BIBentryALTinterwordspacing
S.~L. Lauritzen, ``\BIBforeignlanguage{en}{The {EM} algorithm for graphical
  association models with missing data},''
  \emph{\BIBforeignlanguage{en}{Computational Statistics \& Data Analysis}},
  vol.~19, no.~2, pp. 191--201, Feb. 1995. [Online]. Available:
  \url{https://www.sciencedirect.com/science/article/pii/0167947393E0056A}
\BIBentrySTDinterwordspacing

\bibitem{cooper_bayesian_1992}
\BIBentryALTinterwordspacing
G.~F. Cooper and E.~Herskovits, ``\BIBforeignlanguage{en}{A {Bayesian} method
  for the induction of probabilistic networks from data},''
  \emph{\BIBforeignlanguage{en}{Machine Learning}}, vol.~9, no.~4, pp.
  309--347, Oct. 1992. [Online]. Available:
  \url{https://doi.org/10.1007/BF00994110}
\BIBentrySTDinterwordspacing

\bibitem{van_daalen_privacy_2023}
F.~van Daalen, L.~Ippel, A.~Dekker, and I.~Bermejo, ``Privacy {Preserving}
  n-{Party} {Scalar} {Product} {Protocol},'' \emph{IEEE Transactions on
  Parallel and Distributed Systems}, vol.~34, no.~4, pp. 1060--1066, Apr. 2023,
  conference Name: IEEE Transactions on Parallel and Distributed Systems.

\bibitem{de_marsico_mobile_2015}
\BIBentryALTinterwordspacing
M.~De~Marsico, M.~Nappi, D.~Riccio, and H.~Wechsler,
  ``\BIBforeignlanguage{en}{Mobile {Iris} {Challenge} {Evaluation}
  ({MICHE})-{I}, biometric iris dataset and protocols},''
  \emph{\BIBforeignlanguage{en}{Pattern Recognition Letters}}, vol.~57, pp.
  17--23, May 2015. [Online]. Available:
  \url{https://www.sciencedirect.com/science/article/pii/S0167865515000574}
\BIBentrySTDinterwordspacing

\bibitem{thabtah_autism_2017}
\BIBentryALTinterwordspacing
F.~Thabtah, ``Autism {Spectrum} {Disorder} {Screening}: {Machine} {Learning}
  {Adaptation} and {DSM}-5 {Fulfillment},'' in \emph{Proceedings of the 1st
  {International} {Conference} on {Medical} and {Health} {Informatics} 2017},
  ser. {ICMHI} '17.\hskip 1em plus 0.5em minus 0.4em\relax New York, NY, USA:
  Association for Computing Machinery, 2017, pp. 1--6. [Online]. Available:
  \url{https://doi.org/10.1145/3107514.3107515}
\BIBentrySTDinterwordspacing

\bibitem{lauritzen_local_1988}
\BIBentryALTinterwordspacing
S.~L. Lauritzen and D.~J. Spiegelhalter, ``\BIBforeignlanguage{en}{Local
  {Computations} with {Probabilities} on {Graphical} {Structures} and {Their}
  {Application} to {Expert} {Systems}},'' \emph{\BIBforeignlanguage{en}{Journal
  of the Royal Statistical Society: Series B (Methodological)}}, vol.~50,
  no.~2, pp. 157--194, 1988, \_eprint:
  https://onlinelibrary.wiley.com/doi/pdf/10.1111/j.2517-6161.1988.tb01721.x.
  [Online]. Available:
  \url{https://onlinelibrary.wiley.com/doi/abs/10.1111/j.2517-6161.1988.tb01721.x}
\BIBentrySTDinterwordspacing

\bibitem{smith_using_1988}
\BIBentryALTinterwordspacing
J.~W. Smith, J.~Everhart, W.~Dickson, W.~Knowler, and R.~Johannes, ``Using the
  {ADAP} {Learning} {Algorithm} to {Forecast} the {Onset} of {Diabetes}
  {Mellitus},'' \emph{Proceedings of the Annual Symposium on Computer
  Application in Medical Care}, pp. 261--265, Nov. 1988. [Online]. Available:
  \url{https://www.ncbi.nlm.nih.gov/pmc/articles/PMC2245318/}
\BIBentrySTDinterwordspacing

\bibitem{beinlich_alarm_1989}
\BIBentryALTinterwordspacing
I.~A. Beinlich, H.~J. Suermondt, R.~M. Chavez, and G.~F. Cooper,
  ``\BIBforeignlanguage{en}{The {ALARM} {Monitoring} {System}: {A} {Case}
  {Study} with two {Probabilistic} {Inference} {Techniques} for {Belief}
  {Networks}},'' \emph{\BIBforeignlanguage{en}{AIME 89}}, pp. 247--256, 1989,
  publisher: Springer, Berlin, Heidelberg. [Online]. Available:
  \url{https://link.springer.com/chapter/10.1007/978-3-642-93437-7_28}
\BIBentrySTDinterwordspacing

\bibitem{schlimmer_concept_1987}
\BIBentryALTinterwordspacing
J.~C. Schlimmer, ``\BIBforeignlanguage{en}{Concept acquisition through
  representational adjustment},'' Jul. 1987. [Online]. Available:
  \url{https://escholarship.org/uc/item/48r6d4z0}
\BIBentrySTDinterwordspacing

\end{thebibliography}
	
	\begin{IEEEbiography}[{\includegraphics[width=1in,height=1.25in,clip,keepaspectratio]{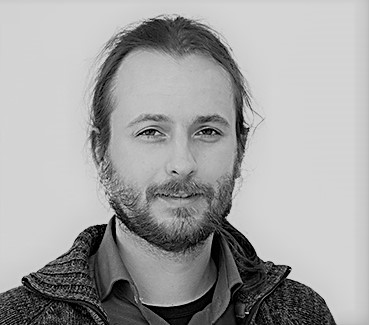}}]{Florian van Daalen}
		Florian van Daalen received his BSc degree in Knowledge Engineering from University Maastricht in 2012 and his MSc degree in Artificial Intelligence in 2014. He is currently working toward the PhD degree in Clinical Data Science within the Clinical Data Science group, University Maastricht, Netherlands. His research interests include privacy preserving techniques, federated learning, and ensemble based learning.
	\end{IEEEbiography}
	
	\begin{IEEEbiography}[{\includegraphics[width=1in,height=1.25in,clip,keepaspectratio]{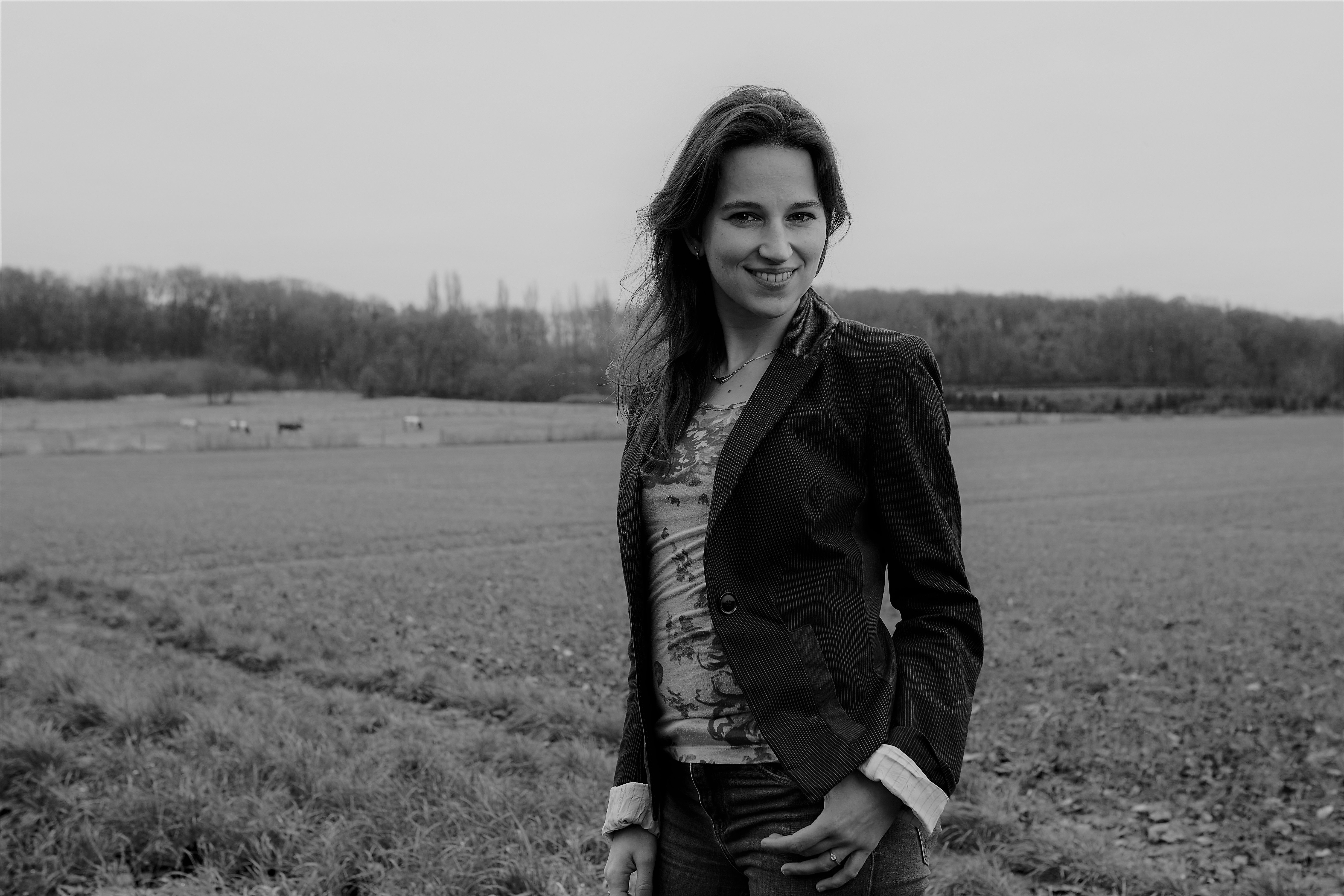}}]{Lianne Ippel}
		Lianne Ippel received her PhD in Statistics from Tilburg University on analyzing data streams with dependent observations, for which she won the dissertation award from General Online Research conference (2018). After a Postdoc at Maastricht University, she now works at Statistics Netherlands where she works at the methodology department on international collaborations and innovative methods for primary data collection.
	\end{IEEEbiography}

	\vspace*{-2\baselineskip}
	
	\begin{IEEEbiography}[{\includegraphics[width=1in,height=1.25in,clip,keepaspectratio]{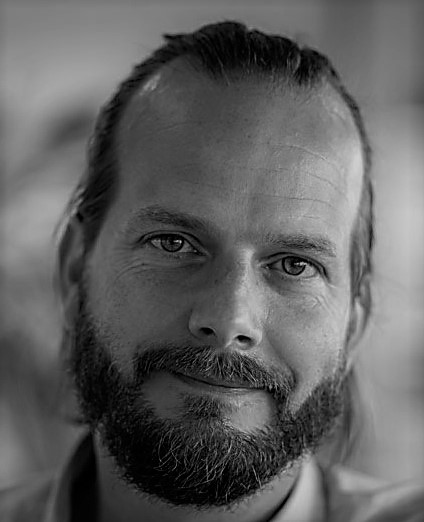}}]{Andre Dekker}
		Prof. Andre Dekker, PhD (1974) is a medical physicist and professor of Clinical Data Science at Maastricht University Medical Center and Maastro Clinic in The Netherlands. His Clinical Data Science research group (50 staff) focuses on 1) federated FAIR data infrastructures, 2) AI for health outcome prediction models and 3) applying AI to improve health. Prof. Dekker has authored over 200 publications, mentored more than 30 PhD students and holds multiple awards and patents on the topic of federated data and AI. He has held visiting scientist appointments at universities and companies in the UK, Australia, Italy, USA and Canada.
	\end{IEEEbiography}
		
	\vspace*{-2\baselineskip}
	\vspace*{-2\baselineskip}
	\vspace*{-2\baselineskip}
	\vspace*{-2\baselineskip}
	\vspace*{-2\baselineskip}
	\vspace*{-2\baselineskip}
	\vspace*{-2\baselineskip}
	\vspace*{-2\baselineskip}
	\vspace*{-2\baselineskip}
	\vspace*{-2\baselineskip}
	
	\begin{IEEEbiography}[{\includegraphics[width=1in,height=1.25in,clip,keepaspectratio]{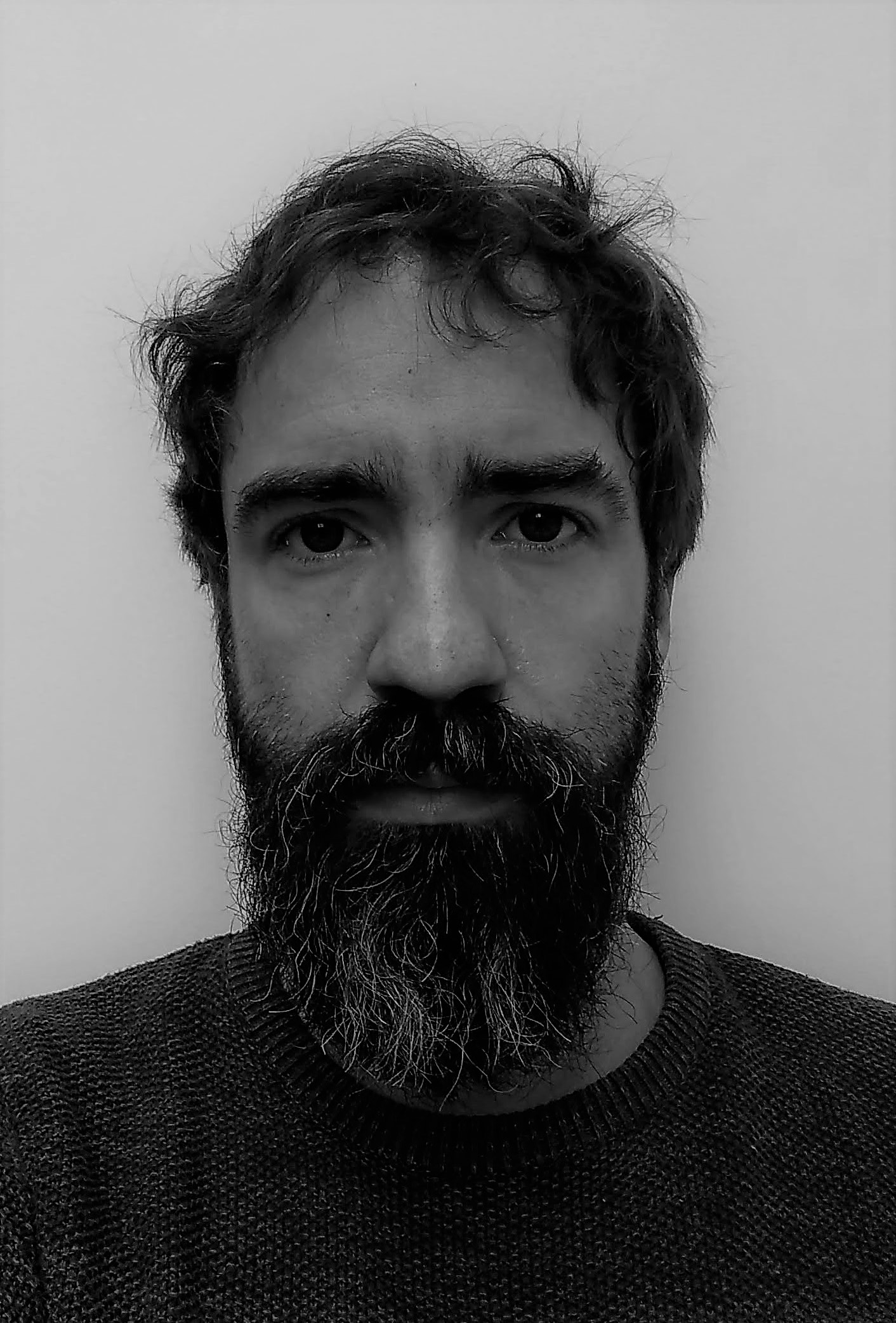}}]{Inigo Bermejo}
		Inigo Bermejo received the BSc degree on Computer Engineering from the University of the Basque Country, Spain, in 2006 and the PhD in Intelligent Systems from UNED, Spain, in 2015. He is currently a postdoctoral researcher at the Clinical Data Science group, Maastricht University. His research interests include privacy preserving techniques, prediction modeling and causal inference. 
	\end{IEEEbiography}	
	
	\appendix
	\label{appendix}
	The remaining experimental results are contained in the following pages of this appendix.
	
\begin{table*}[t]
	\centering
	\caption{Experimental results vertically split 3-party scenarios where attributes were randomly split across parties. '*' Indicates the best performing model, '\dag' indicates the second best performing model.}
	\begin{tabular}{c|c|c|c|c|c|c|c}
		\multicolumn{2}{c}{} \vline & \multicolumn{6}{c}{AUC} \\ \hline													
		Name	&	Missing Data Level	&	FBNE	&	Party 1	&	Party 2	 & Party 3 &	Central	&	VertiBayes	\\ \hline
		\makecell{Alarm \\ population size: 10000}	&	0	&	0,884*	&	0,790	&	0,669	&	0,561	&	0,790\dag	&	0,790\dag	\\ \hline
		\multirow{4}{*}{\begin{tabular}{c}Asia \\ population size: 10000 \end{tabular}}	&	0	&	0,997*	&	0,824	&	0,873	&	0,902	&	0,996\dag	&	0,986	\\
		&	0.05	&	0,739\dag	&	0,657	&	0,651	&	0,664	&	0,736	&	0,750*	\\
		&	0.1	&	0,622\dag	&	0,607	&	0,519	&	0,583	&	0,621	&	0,704*	\\
		&	0.3	&	0,418\dag	&	0,380	&	0,394	&	0,407	&	0,417	&	0,569*	\\ \hline
		\multirow{4}{*}{\begin{tabular}{c}Autism \\ population size: 704 \end{tabular}}	&	0	&	0,934\dag	&	0,784	&	0,787	&	0,740	&	0,832	&	0,977*	\\
		&	0.05	&	0,797*	&	0,742	&	0,664	&	0,434	&	0,732	&	0,780\dag	\\
		&	0.1	&	0,734*	&	0,686	&	0,605	&	0,447	&	0,694	&	0,730\dag	\\
		&	0.3	&	0,497\dag	&	0,492	&	0,388	&	0,369	&	0,489	&	0,627*	\\ \hline
		\multirow{4}{*}{\begin{tabular}{c}Diabetes \\ population size: 768 \end{tabular}}	&	0	&	0,801*	&	0,659	&	0,647	&	0,675	&	0,789\dag	&	0,783	\\
		&	0.05	&	0,753*	&	0,654	&	0,630	&	0,598	&	0,728	&	0,740\dag	\\
		&	0.1	&	0,695\dag	&	0,613	&	0,614	&	0,560	&	0,678	&	0,804*	\\
		&	0.3	&	0,445\dag	&	0,407	&	0,383	&	0,382	&	0,438	&	0,552*	\\ \hline
		\makecell{Mushroom \\ population size: 8124}	&	0	&	0,989*	&	0,818	&	0,987	&	0,589	&	0,988\dag	&	0,987	\\ \hline
	\end{tabular}
\end{table*}

\begin{table*}[t]
	\centering
	\caption{Experimental results vertically split 3-party scenarios where attributes were manually split across parties. '*' Indicates the best performing model, '\dag' indicates the second best performing model.}
	\begin{tabular}{c|c|c|c|c|c|c|c}
		\multicolumn{2}{c}{} \vline & \multicolumn{6}{c}{AUC} \\ \hline													
		Name	&	Missing Data Level	&	FBNE	&	Party 1	&	Party 2	 & Party 3 &	Central	&	VertiBayes	\\ \hline
		\multirow{4}{*}{\begin{tabular}{c}Autism \\ population size: 704 \end{tabular}}	&	0	&	0,920*	&	0,843	&	0,730	&	0,811	&	0,830\dag	&	0,829	\\
		&	0.05	&	0,797*	&	0,742	&	0,664	&	0,434	&	0,732	&	0,780\dag	\\
		&	0.1	&	0,734*	&	0,686	&	0,605	&	0,447	&	0,694	&	0,730\dag	\\
		&	0.3	&	0,497\dag	&	0,492	&	0,388	&	0,369	&	0,489	&	0,627*	\\ \hline
		\makecell{Mushroom \\ population size: 8124}	&	0	&	0,991*	&	0,881	&	0,986	&	0,680	&	0,988\dag	&	0,986	\\ \hline
	\end{tabular}
\end{table*}

\begin{table*}[t]
	\centering
	\caption{Experimental results hybrid split 3-party scenarios where hybrid split attributes can fully incorperated into the local models. '*' Indicates the best performing model, '\dag' indicates the second best performing model.}
	\begin{tabular}{c|c|c|c|c|c|c|c}
		\multicolumn{2}{c}{} \vline & \multicolumn{6}{c}{AUC} \\ \hline													
		Name	&	Missing Data Level	&	FBNE	&	Party 1	&	Party 2	 & Party 3 &	Central	&	VertiBayes	\\ \hline
		\multirow{4}{*}{\begin{tabular}{c}Asia \\ population size: 10000 \end{tabular}}	&	0	&	0,996\dag	&	0,885	&	0,929	&	0,929	&	0,995\dag	&	0,999*	\\
		&	0.05	&	0,743	&	0,709	&	0,717	&	0,722	&	0,745\dag	&	0,766*	\\
		&	0.1	&	0,623\dag	&	0,612	&	0,554	&	0,555	&	0,619	&	0,669*	\\
		&	0.3	&	0,419\dag	&	0,401	&	0,410	&	0,412	&	0,419\dag	&	0,567*	\\ \hline
		\multirow{4}{*}{\begin{tabular}{c}Autism \\ population size: 704 \end{tabular}}	&	0	&	0,903*	&	0,807	&	0,817	&	0,822	&	0,849\dag	&	0,847	\\
		&	0.05	&	0,793*	&	0,724	&	0,710	&	0,715	&	0,753	&	0,777\dag	\\
		&	0.1	&	0,740\dag	&	0,671	&	0,667	&	0,671	&	0,682	&	0,749*	\\
		&	0.3	&	0,527\dag	&	0,481	&	0,493	&	0,493	&	0,503	&	0,747*	\\ \hline
		\multirow{4}{*}{\begin{tabular}{c}Diabetes \\ population size: 768 \end{tabular}}	&	0	&	0,811*	&	0,731	&	0,695	&	0,700	&	0,779\dag	&	0,776	\\
		&	0.05	&	0,692\dag	&	0,604	&	0,608	&	0,607	&	0,673	&	0,734*	\\
		&	0.1	&	0,755*	&	0,670	&	0,670	&	0,673	&	0,726	&	0,752\dag	\\
		&	0.3	&	0,456\dag	&	0,416	&	0,403	&	0,404	&	0,439	&	0,697*	\\ \hline
		\multirow{4}{*}{\begin{tabular}{c}Iris \\ population size: 150 \end{tabular}}	&	0	&	0,939*	&	0,889\dag	&	0,879	&	0,886	&	0,883	&	0,771	\\
		&	0.05	&	0,892*	&	0,783	&	0,835	&	0,830	&	0,876\dag	&	0,736	\\
		&	0.1	&	0,788*	&	0,702	&	0,719	&	0,724\dag	&	0,713	&	0,704	\\
		&	0.3	&	0,653\dag	&	0,588	&	0,595	&	0,599	&	0,662*	&	0,611	\\ \hline			
	\end{tabular}
\end{table*}

\begin{table*}[t]
	\centering
	\caption{Experimental results horizontally split 3-party scenarios where records are randomly split across parties. '*' Indicates the best performing model, '\dag' indicates the second best performing model.}
	\begin{tabular}{c|c|c|c|c|c|c|c|c}
		\multicolumn{2}{c}{} \vline & \multicolumn{6}{c}{AUC} \\ \hline													
		Name	&	Missing Data Level	&	FBNE	&	Party 1	&	Party 2	 & Party 3 &	Central	&	VertiBayes	\\ \hline
		\multirow{4}{*}{\begin{tabular}{c}Asia \\ population size: 10000 \end{tabular}}	&	0	&	0,995*	&	0,995*	&	0,995*	&	0,995*	&	0,995*	&	0,987	\\
		&	0.05	&	0,741\dag	&	0,741\dag	&	0,741\dag	&	0,741\dag	&	0,730	&	0,763*	\\
		&	0.1	&	0,623\dag	&	0,623\dag	&	0,623\dag	&	0,623\dag	&	0,618	&	0,674*	\\
		&	0.3	&	0,418\dag	&	0,418\dag	&	0,418\dag	&	0,418\dag	&	0,417	&	0,568*	\\ \hline
		\multirow{4}{*}{\begin{tabular}{c}Autism \\ population size: 704 \end{tabular}}	&	0	&	0,889*	&	0,836	&	0,836	&	0,836	&	0,838\dag	&	0,829	\\
		&	0.05	&	0,794*	&	0,736	&	0,736	&	0,736	&	0,754	&	0,780\dag	\\
		&	0.1	&	0,724\dag	&	0,687	&	0,687	&	0,687	&	0,688	&	0,749*	\\
		&	0.3	&	0,544\dag	&	0,493	&	0,494	&	0,493	&	0,494	&	0,625*	\\ \hline
		\multirow{4}{*}{\begin{tabular}{c}Diabetes \\ population size: 768 \end{tabular}}	&	0	&	0,775	&	0,780\dag	&	0,780\dag	&	0,780\dag	&	0,786*	&	0,778	\\
		&	0.05	&	0,730\dag	&	0,727	&	0,727	&	0,727	&	0,720	&	0,753*	\\
		&	0.1	&	0,674\dag	&	0,673	&	0,673	&	0,673	&	0,667	&	0,733*	\\
		&	0.3	&	0,448\dag	&	0,439	&	0,437	&	0,438	&	0,439	&	0,648*	\\ \hline
		\multirow{4}{*}{\begin{tabular}{c}Iris \\ population size: 150 \end{tabular}}	&	0	&	0,960*	&	0,896	&	0,897\dag	&	0,890	&	0,890	&	0,761	\\
		&	0.05	&	0,875	&	0,876	&	0,877\dag	&	0,875	&	0,879*	&	0,768	\\
		&	0.1	&	0,826*	&	0,703	&	0,703	&	0,703	&	0,709	&	0,676	\\
		&	0.3	&	0,632	&	0,666*	&	0,666*	&	0,666*	&	0,664\dag	&	0,626	\\ \hline		
	\end{tabular}
\end{table*}

\begin{table*}[t]
	\centering
	\caption{Experimental results horizontally split 2-party scenarios where records are randomly split across parties. Varies levels of bias were introduced in this experiment where the level of bias corresponds to the probability of an individual with first class label to be assigned to party 1. '*' Indicates the best performing model, '\dag' indicates the second best performing model.}
	\begin{tabular}{c|c|c|c|c|c|c|c}
		\multicolumn{3}{c}{} \vline & \multicolumn{5}{c}{AUC} \\ \hline													
		Name	&	Bias Level	&	Missing Data Level	& Ensemble	&	 Party 1	&	 Part 2	&	 Central & 	 VertiBayes	\\ \hline
		\multirow{12}{*}{\begin{tabular}{c}Asia \\ population size: 10000 \end{tabular}}	&	\multirow{4}{*}{\begin{tabular}{c} 0.75 \end{tabular}}	&	0	&	0,995\dag	&	0,995\dag	&	0,995\dag	&	0,996*	&	0,987	\\
		&		&	0.05	&	0,743\dag	&	0,741	&	0,741	&	0,742	&	0,765*	\\
		&		&	0.1	&	0,624\dag	&	0,623	&	0,623	&	0,624\dag	&	0,670*	\\
		&		&	0.3	&	0,419\dag	&	0,419\dag	&	0,418	&	0,417	&	0,568*	\\ \cline{2-8}
		&	\multirow{4}{*}{\begin{tabular}{c} 0.85 \end{tabular}}	&	0	&	0,996*	&	0,995\dag	&	0,995\dag	&	0,996*	&	0,986	\\
		&		&	0.05	&	0,741\dag	&	0,741\dag	&	0,741\dag	&	0,737	&	0,763*	\\
		&		&	0.1	&	0,622	&	0,623	&	0,623	&	0,628\dag	&	0,671*	\\
		&		&	0.3	&	0,419	&	0,419	&	0,419	&	0,420\dag	&	0,568*	\\ \cline{2-8}
		&	\multirow{4}{*}{\begin{tabular}{c} 0.95 \end{tabular}}	&	0	&	0,995\dag	&	0,995\dag	&	0,995\dag	&	0,996*	&	0,986	\\
		&		&	0.05	&	0,741	&	0,741	&	0,741	&	0,742\dag	&	0,764*	\\
		&		&	0.1	&	0,624	&	0,623	&	0,623	&	0,627\dag	&	0,669*	\\
		&		&	0.3	&	0,420\dag	&	0,419	&	0,419	&	0,418	&	0,567*	\\ \hline
		\multirow{12}{*}{\begin{tabular}{c}Autism \\ population size: 704 \end{tabular}}	&	\multirow{4}{*}{\begin{tabular}{c} 0.75 \end{tabular}}	&	0	&	0,876*	&	0,500	&	0,780	&	0,845\dag	&	0,835	\\
		&		&	0.05	&	0,786*	&	0,488	&	0,487	&	0,748	&	0,780\dag	\\
		&		&	0.1	&	0,708\dag	&	0,501	&	0,487	&	0,698	&	0,749*	\\
		&		&	0.3	&	0,535\dag	&	0,380	&	0,456	&	0,500	&	0,627*	\\ \cline{2-8}
		&	\multirow{4}{*}{\begin{tabular}{c} 0.85 \end{tabular}}	&	0	&	0,880*	&	0,500	&	0,770	&	0,848\dag	&	0,835	\\
		&		&	0.05	&	0,779*	&	0,464	&	0,464	&	0,731\dag	&	0,779*	\\
		&		&	0.1	&	0,722\dag	&	0,445	&	0,450	&	0,688	&	0,746*	\\
		&		&	0.3	&	0,527\dag	&	0,427	&	0,454	&	0,493	&	0,630*	\\ \cline{2-8}
		&	\multirow{4}{*}{\begin{tabular}{c} 0.95 \end{tabular}}	&	0	&	0,898*	&	0,534	&	0,500	&	0,830	&	0,834\dag	\\
		&		&	0.05	&	0,779*	&	0,464	&	0,512	&	0,736\dag	&	0,779*	\\
		&		&	0.1	&	0,724\dag	&	0,443	&	0,528	&	0,675	&	0,747*	\\
		&		&	0.3	&	0,512\dag	&	0,430	&	0,450	&	0,499	&	0,629*	\\ \hline
		\multirow{12}{*}{\begin{tabular}{c}Diabetes \\ population size: 768 \end{tabular}}	&	\multirow{4}{*}{\begin{tabular}{c} 0.75 \end{tabular}}	&	0	&	0,778	&	0,500	&	0,500	&	0,787*	&	0,780\dag	\\
		&		&	0.05	&	0,737\dag	&	0,480	&	0,480	&	0,731	&	0,753*	\\
		&		&	0.1	&	0,676\dag	&	0,447	&	0,447	&	0,674	&	0,736*	\\
		&		&	0.3	&	0,430	&	0,405	&	0,359	&	0,445\dag	&	0,645*	\\ \cline{2-8}
		&	\multirow{4}{*}{\begin{tabular}{c} 0.85 \end{tabular}}	&	0	&	0,781*	&	0,500	&	0,500	&	0,776\dag	&	0,781*	\\
		&		&	0.05	&	0,730\dag	&	0,480	&	0,480	&	0,730\dag	&	0,754*	\\
		&		&	0.1	&	0,675\dag	&	0,447	&	0,447	&	0,678	&	0,736*	\\
		&		&	0.3	&	0,412	&	0,349	&	0,406	&	0,444\dag	&	0,645*	\\ \cline{2-8}
		&	\multirow{4}{*}{\begin{tabular}{c} 0.95 \end{tabular}}	&	0	&	0,772	&	0,500	&	0,500	&	0,786*	&	0,780\dag	\\
		&		&	0.05	&	0,581	&	0,480	&	0,499	&	0,732\dag	&	0,754*	\\
		&		&	0.1	&	0,530	&	0,447	&	0,459	&	0,672\dag	&	0,737*	\\
		&		&	0.3	&	0,354	&	0,370	&	0,346	&	0,442\dag	&	0,646*	\\ \hline
		\multirow{12}{*}{\begin{tabular}{c}Iris \\ population size: 150 \end{tabular}}	&	\multirow{4}{*}{\begin{tabular}{c} 0.75 \end{tabular}}	&	0	&	0,942*	&	0,876	&	0,735	&	0,890\dag	&	0,767	\\
		&		&	0.05	&	0,870\dag	&	0,727	&	0,782	&	0,875*	&	0,760	\\
		&		&	0.1	&	0,802*	&	0,588	&	0,703	&	0,718\dag	&	0,669	\\
		&		&	0.3	&	0,608	&	0,544	&	0,611	&	0,676*	&	0,607	\\ \cline{2-8}
		&	\multirow{4}{*}{\begin{tabular}{c} 0.85 \end{tabular}}	&	0	&	0,950*	&	0,782	&	0,692	&	0,896\dag	&	0,779	\\
		&		&	0.05	&	0,870\dag	&	0,625	&	0,793	&	0,879*	&	0,766	\\
		&		&	0.1	&	0,746*	&	0,604	&	0,712	&	0,710	&	0,678	\\
		&		&	0.3	&	0,636\dag	&	0,478	&	0,571	&	0,669*	&	0,600	\\ \cline{2-8}
		&	\multirow{4}{*}{\begin{tabular}{c} 0.95 \end{tabular}}	&	0	&	0,915*	&	0,560	&	0,653	&	0,895\dag	&	0,780	\\
		&		&	0.05	&	0,800\dag	&	0,497	&	0,688	&	0,873*	&	0,771	\\
		&		&	0.1	&	0,752*	&	0,504	&	0,641	&	0,710\dag	&	0,667	\\
		&		&	0.3	&	0,633\dag	&	0,469	&	0,480	&	0,673*	&	0,608	\\ \hline		 		
	\end{tabular}
\end{table*}	

\begin{table*}[t]
	\centering
	\caption{Experimental results horizontally split 3-party scenarios where records are randomly split across parties. Varies levels of bias were introduced in this experiment where the level of biass corresponds to the probability of an individual with first class label to be assigned to party 1. '*' Indicates the best performing model, '\dag' indicates the second best performing model.}
	\begin{tabular}{c|c|c|c|c|c|c|c|c}
		\multicolumn{3}{c}{} \vline & \multicolumn{6}{c}{AUC} \\ \hline													
		Name	&	Bias Level	&	Missing Data Level	& Ensemble	&	 Party 1	&	 Party 2 & Party 3 & Central & 	 VertiBayes	\\ \hline
		\multirow{12}{*}{\begin{tabular}{c}Asia \\ population size: 10000 \end{tabular}}	&	\multirow{4}{*}{\begin{tabular}{c} 0.75 \end{tabular}}	&	0	&	0,996*	&	0,995\dag	&	0,995\dag	&	0,995\dag	&	0,995\dag	&	0,988	\\
		&		&	0.05	&	0,740	&	0,741	&	0,741	&	0,741	&	0,745\dag	&	0,766*	\\
		&		&	0.1	&	0,622	&	0,623	&	0,623	&	0,623	&	0,624\dag	&	0,668*	\\
		&		&	0.3	&	0,419\dag	&	0,418	&	0,418	&	0,418	&	0,418	&	0,569*	\\ \cline{2-9}
		&	\multirow{4}{*}{\begin{tabular}{c} 0.85 \end{tabular}}	&	0	&	0,996*	&	0,995\dag	&	0,995\dag	&	0,995\dag	&	0,996*	&	0,986	\\
		&		&	0.05	&	0,741\dag	&	0,741\dag	&	0,741\dag	&	0,741\dag	&	0,740	&	0,769*	\\
		&		&	0.1	&	0,625	&	0,623	&	0,623	&	0,623	&	0,628\dag	&	0,671*	\\
		&		&	0.3	&	0,419\dag	&	0,418	&	0,418	&	0,418	&	0,417	&	0,568*	\\ \cline{2-9}
		&	\multirow{4}{*}{\begin{tabular}{c} 0.95 \end{tabular}}	&	0	&	0,995\dag	&	0,995\dag	&	0,995\dag	&	0,995\dag	&	0,996*	&	0,987	\\
		&		&	0.05	&	0,743	&	0,741	&	0,741	&	0,741	&	0,745\dag	&	0,762*	\\
		&		&	0.1	&	0,622	&	0,623	&	0,623	&	0,623	&	0,623\dag	&	0,671*	\\
		&		&	0.3	&	0,419\dag	&	0,418	&	0,418	&	0,418	&	0,416	&	0,569*	\\ \hline
		\multirow{12}{*}{\begin{tabular}{c}Autism \\ population size: 704 \end{tabular}}	&	\multirow{4}{*}{\begin{tabular}{c} 0.75 \end{tabular}}	&	0	&	0,911*	&	0,836	&	0,836	&	0,836	&	0,843\dag	&	0,833	\\
		&		&	0.05	&	0,797*	&	0,736	&	0,736	&	0,736	&	0,732	&	0,776\dag	\\
		&		&	0.1	&	0,728\dag	&	0,687	&	0,687	&	0,687	&	0,700	&	0,746*	\\
		&		&	0.3	&	0,541\dag	&	0,493	&	0,494	&	0,494	&	0,491	&	0,627*	\\ \cline{2-9}
		&	\multirow{4}{*}{\begin{tabular}{c} 0.85 \end{tabular}}	&	0	&	0,905*	&	0,836	&	0,836	&	0,836	&	0,832	&	0,842\dag	\\
		&		&	0.05	&	0,810*	&	0,736	&	0,736	&	0,736	&	0,732	&	0,785\dag	\\
		&		&	0.1	&	0,739\dag	&	0,687	&	0,688	&	0,687	&	0,687	&	0,744*	\\
		&		&	0.3	&	0,541\dag	&	0,493	&	0,494	&	0,494	&	0,492	&	0,623*	\\ \cline{2-9}
		&	\multirow{4}{*}{\begin{tabular}{c} 0.95 \end{tabular}}	&	0	&	0,899*	&	0,836	&	0,836	&	0,836	&	0,844\dag	&	0,835	\\
		&		&	0.05	&	0,785*	&	0,736	&	0,736	&	0,736	&	0,745	&	0,780\dag	\\
		&		&	0.1	&	0,720\dag	&	0,687	&	0,687	&	0,687	&	0,677	&	0,745*	\\
		&		&	0.3	&	0,490	&	0,494	&	0,494	&	0,494	&	0,495\dag	&	0,630*	\\ \hline
		\multirow{12}{*}{\begin{tabular}{c}Diabetes \\ population size: 768 \end{tabular}}	&	\multirow{4}{*}{\begin{tabular}{c} 0.75 \end{tabular}}	&	0	&	0,788*	&	0,780	&	0,780	&	0,780	&	0,779	&	0,786\dag	\\
		&		&	0.05	&	0,760\dag	&	0,727	&	0,727	&	0,727	&	0,727	&	0,761*	\\
		&		&	0.1	&	0,690\dag	&	0,673	&	0,672	&	0,672	&	0,680	&	0,743*	\\
		&		&	0.3	&	0,430	&	0,437	&	0,438\dag	&	0,438\dag	&	0,434	&	0,653*	\\ \cline{2-9}
		&	\multirow{4}{*}{\begin{tabular}{c} 0.85 \end{tabular}}	&	0	&	0,785*	&	0,780\dag	&	0,780\dag	&	0,780\dag	&	0,776	&	0,777	\\
		&		&	0.05	&	0,745\dag	&	0,727	&	0,727	&	0,727	&	0,726	&	0,747*	\\
		&		&	0.1	&	0,674	&	0,673	&	0,673	&	0,673	&	0,676\dag	&	0,737*	\\
		&		&	0.3	&	0,394	&	0,437	&	0,438	&	0,439\dag	&	0,438	&	0,648*	\\ \cline{2-9}
		&	\multirow{4}{*}{\begin{tabular}{c} 0.95 \end{tabular}}	&	0	&	0,752	&	0,780\dag	&	0,780\dag	&	0,780\dag	&	0,782*	&	0,780\dag	\\
		&		&	0.05	&	0,631	&	0,727	&	0,727	&	0,727	&	0,723\dag	&	0,755*	\\
		&		&	0.1	&	0,564	&	0,672	&	0,673	&	0,673\dag	&	0,668	&	0,740*	\\
		&		&	0.3	&	0,353	&	0,438	&	0,438	&	0,438	&	0,443\dag	&	0,649*	\\ \hline				 		
	\end{tabular}
\end{table*}

\end{document}